\documentclass[sigconf]{acmart}

\usepackage{caption}
\usepackage{multirow}
\usepackage{paralist}
\usepackage{booktabs}

\AtBeginDocument{%
  \providecommand\BibTeX{{%
    \normalfont B\kern-0.5em{\scshape i\kern-0.25em b}\kern-0.8em\TeX}}}

\copyrightyear{2021}
\acmYear{2021}
\setcopyright{rightsretained}
\acmConference[CIKM '21]{Proceedings of the 30th ACM International
Conference on Information and Knowledge Management}{November 1--5,
2021}{Virtual Event, QLD, Australia}
\acmBooktitle{Proceedings of the 30th ACM International Conference on
Information and Knowledge Management (CIKM '21), November 1--5,
2021, Virtual Event, QLD, Australia}
\acmDOI{10.1145/3459637.3481916}
\acmISBN{978-1-4503-8446-9/21/11}

\begin{document}

\fancyhead{}

\title{ETA Prediction with Graph Neural Networks in Google Maps}

\author{Austin Derrow-Pinion$^1$, Jennifer She$^1$, David Wong$^{2*}$, Oliver Lange$^3$,  Todd Hester$^{4*}$, Luis Perez$^{5*}$, Marc Nunkesser$^3$, Seongjae Lee$^3$, Xueying Guo$^3$, Brett Wiltshire$^1$, Peter W. Battaglia$^1$, Vishal Gupta$^1$, Ang Li$^1$, Zhongwen Xu$^{6*}$, Alvaro Sanchez-Gonzalez$^1$, Yujia Li$^1$ and Petar Veli\v{c}kovi\'{c}$^1$}
\affiliation{%
  \institution{$^1$DeepMind \quad $^2$Waymo \quad $^3$Google \quad $^4$Amazon \quad $^5$Facebook AI \quad $^6$Sea AI Lab \quad $^*$work done while at DeepMind}
  \country{}
  \city{}
}
\email{{derrowap,jenshe,wongda,petarv}@google.com}

%
\renewcommand{\shortauthors}{Derrow-Pinion, She, Wong, et al.}

\begin{abstract}

Travel-time prediction constitutes a task of high importance in transportation networks, with web mapping services like Google Maps regularly serving vast quantities of travel time queries from users and enterprises alike. Further, such a task requires accounting for complex spatiotemporal interactions (modelling both the topological properties of the road network and anticipating events---such as rush hours---that may occur in the future). Hence, it is an ideal target for graph representation learning at scale. Here we present a graph neural network estimator for estimated time of arrival (ETA) which we have deployed in production at Google Maps.
While our main architecture consists of standard GNN building blocks, we further detail the usage of training schedule methods such as MetaGradients in order to make our model robust and production-ready. We also provide prescriptive studies: ablating on various architectural decisions and training regimes, and qualitative analyses on real-world situations where our model provides a competitive edge. Our GNN proved powerful when deployed, significantly reducing negative ETA outcomes in several regions compared to the previous production baseline (40+\% in cities like Sydney).
\end{abstract}




\maketitle

\section{Introduction}
\begin{figure}[t]
    \centering
    \includegraphics[width=\linewidth]{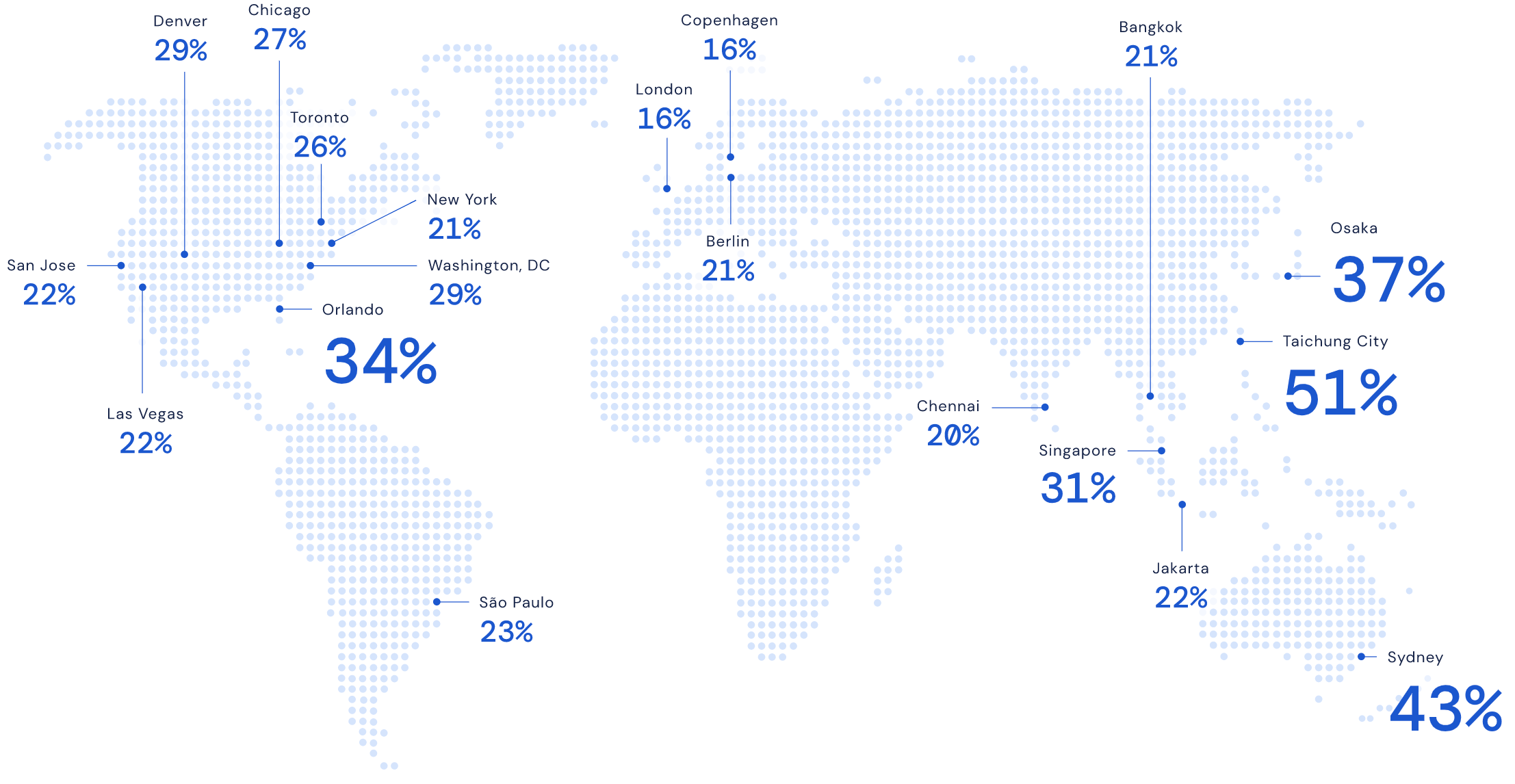}
    \caption{Google Maps estimated time-of-arrival (ETA) prediction improvements for several world regions, when using our deployed graph neural network-based estimator. Numbers represent relative reduction in negative ETA outcomes compared to the prior approach used in production. A negative ETA outcome occurs when the ETA error from the observed travel duration is over some threshold and acts as a measure of accuracy.}
    \label{fig:maps_impro}
\end{figure}
Web mapping services such as Google Maps are invaluable tools for many users for interactively navigating areas of the Earth---especially so in metropolitan areas. On a daily level, such products are used not only by everyday users finding the (optimal) routes between waypoints, but also by enterprises which rely on accurate mapping features (such as food delivery services, which may seek to provide accurate and optimised delivery time estimates). One critical task in this context is \emph{expected time of arrival} (ETA) prediction: given current conditions on the road network, and for a particular candidate route between waypoints, predict the expected time of travel along this route.

The significance of such a predictor is clear: accurate ETA predictions allow traffic participants to make more informed decisions, potentially avoiding congested areas and minimising overall time spent in traffic. It is important to note that powerful ETA predictors need to meaningfully reason about conditions which take place in the \emph{future}, and may not be directly obvious from current road state. As a simple example, the number of traffic participants may substantially increase during daily \emph{rush-hour} periods, when most people commute to and from the workplace---but in many cases such situations may be more subtle. Further, there is a wealth of historical travel data available which may be used to support detection and exploitation on such subtleties. As such, this problem is substantially amenable to machine learning approaches.

As the road network is naturally modelled by a \emph{graph} of road segments and intersections, ETA prediction is amenable to graph representation learning \cite{bronstein2017geometric,battaglia2018relational,hamilton2020graph} approaches, particularly \emph{graph neural networks} (GNNs) \cite{kipf2016semi,velivckovic2017graph,gilmer2017neural}. Here we present our graph neural network model for ETA prediction, which we deployed in production at Google Maps, observing significant reductions in negative ETA outcomes across all trips worldwide---above $40\%$ in cities like Sydney---compared to the previous production baseline. See Figure \ref{fig:maps_impro} for improvements in additional regions. A negative ETA outcome occurs when the ETA error from the observed travel duration is over some threshold and acts as a measure of accuracy.

The task of arrival time estimation can be described as providing an estimated travel time given a user-provided starting location and a route suggested by a prior system. This problem is difficult because it needs to take into account both spatial information (captured within the road network) as well as temporal information (the evolution of traffic conditions over time). In particular, we need to anticipate that travel times across road segments may be affected by traffic conditions that are further away, or not immediately on the user track---potentially also accounting for ``background'' signals such as traffic light configurations. We also need to anticipate that traffic conditions may have changed by the time a user traverses to a road segment further along their route.

\paragraph{Contributions} We demonstrate how a powerful graph neural network estimator can be constructed for the ETA prediction task, while following simple design principles and carefully designed loss functions. Such a GNN then remains stable and performant when deployed in production, providing substantial gains against prior production baselines which did not use the road network graph reprsentations. Beyond this, our key contributions to this problem are as follows:
\begin{itemize}
    \item Innovations on how the road network data is \textbf{featurised} and presented to the GNN;
    \item Innovations in the GNN \textbf{model design} and \textbf{operation}. While the GNN itself is constructed of standard building blocks, we found several benefits for the training regime (such as MetaGradients \cite{xu2018meta} and semi-supervised training \cite{velickovic2019deep,kipf2016variational}) as well as architectural ablations (e.g. carefully tuned combinations of aggregators \cite{corso2020principal}) that proved particularly important for maintaining stability of the GNN in this regime and making it production-ready.
    \item Finally, we \textbf{deploy} the trained GNN model in production within Google Maps, observing strong \textbf{real-world benefits}. As displayed in Figure \ref{fig:maps_impro}, we noticed clear quantitative improvements in negative ETA outcomes, but besides the on-line performance, we performed several offline ablations on various architectural and setup decisions, as well as visualisations of particular traffic situations where our model has a clear advantage over the prior baseline.
\end{itemize}

On one hand, our work represents another in a series of successful \emph{web-scale} deployed applications of graph neural networks with real-time positive user experience impact \cite{ying2018graph,yang2019aligraph,pal2020pinnersage,hao2020p}. On another, we offer ablative studies and insights into the GNN model's predictions which simultaneously \begin{inparaenum}
\item[(a)] elucidate the kinds of real-world traffic use cases where such a model may provide a significant edge; and
\item[(b)] provide prescriptive advice for other practitioners in the general area of user-facing traffic recommendation.
\end{inparaenum}

\section{Related Work}
We outline several related works that either directly inspire the building blocks of our proposed GNN architecture and pipeline, or sit more broadly in the space of GNNs for traffic forecasting. 

\paragraph{Graph representation learning}

Being a graph neural network, our work directly builds upon the previous years of progress in graph representation learning. In particular, our model is fundamentally based on the Graph Network framework \cite{battaglia2018relational} and follows the encode-process-decode \cite{hamrick2018relational} paradigm. This allows us to align better with the iterative nature of traffic computations, as well as pathfinding algorithms. Improving such an alignment is known to help GNNs generalise better to a diverse distribution of graphs \cite{velivckovic2019neural}.

\paragraph{Travel-time prediction} Given its wide applicability, travel-time prediction is a problem that has historically been closely studied under various settings. Earlier work modifies convolutional neural networks to be mindful of the spatial properties of the trajectory \cite{wang2018will} and employs graph neural networks coupled with recurrent mechanisms \cite{hu2020stochastic}. The concurrently developed work of Yuan \emph{et al.} \cite{yuan2020effective} makes an important observation in this space: travel time is often largely dependent on the historical travel times in the given time-of-day---an observation that we extensively use. The recent CurbGAN framework \cite{zhang2020curb} may also be of interest, with its leveraging of generative adversarial networks for evaluating urban development plans in the context of estimated travel times.

\paragraph{Spatiotemporal traffic forecasting}

Using GNNs to forecast traffic conditions is historically a very active direction: most influential papers in the area of \emph{spatiotemporal graph representation learning} rely on traffic datasets. The spatiotemporal setup typically assumes a graph structure over which each node maintains a \emph{time series}---in the case of road networks, such time series typically correspond to historical speeds measured at regular time intervals. While our GNN model detailed here processes \emph{static} inputs, and does not directly forecast future speeds, having some estimate about future traffic flow behaviour is likely to be meaningful for ETA prediction.

One of the first influential papers in the area was the \emph{diffusion-convolutional recurrent neural network} (DCRNN) \cite{li2017diffusion}. This paper illustrated how explicitly accounting for the graph structure provides a significant reduction in forecasting error over several horizons. This generic blueprint of a spatial GNN component combined with a temporal component yielded several subsequent proposals, including STGCN \cite{yu2017spatio}, GaAN \cite{zhang2018gaan}, Graph WaveNet \cite{wu2019graph}, ST-GRAT \cite{park2020st}, StemGNN \cite{cao2020stem}, and GMAN \cite{zheng2020gman}. We refer interested readers to \cite{shi2018machine} for a survey of interesting problems and methodologies in the area.

Lastly, we find it important to mention that the assumption of the time-series being aligned across nodes does not necessarily hold in practice---we may instead expect data to be provided in an \emph{asynchronous} fashion. A recent line of research on \emph{dynamic graph representation learning} \cite{xu2020inductive,rossi2020temporal} tries to explicitly account for such situations, both on the featurisation and the topological side.

\paragraph{Graph neural networks at scale}

As graph neural network methodologies became more widely available and scalable \cite{hamilton2017inductive,chen2018fastgcn,rossi2020sign}, this also enabled a wider range of \emph{web-scale} applications of graph representation learning. Perhaps the most popularised applications concerned recommender systems, which are very naturally representable as a graph-structured task: with Pinterest being one of the most early adopters \cite{ying2018graph,pal2020pinnersage}. GNNs have also been deployed for product recommendation at Amazon \cite{hao2020p}, E-commerce applications at Alibaba \cite{yang2019aligraph}, engagement forecasting and friend ranking in Snapchat \cite{tang2020knowing,sankar2021graph}, and most relevantly, they are powering traffic predictions within Baidu Maps \cite{fang2020constgat}.

\paragraph{Training regimes and components}

While the core components of our architecture correspond to the Graph Network paradigm, ETA prediction in production invites particularly unstable training conditions across many batches of queries, particularly over routes of different scales. We adapted the MetaGradients \cite{xu2018meta} methodology from reinforcement learning, which allowed us to dynamically tune the learning rate during training and stabilise it across many uneven query batches, enabling a production-ready GNN model.

Further, there exist many patterns inherent to the \emph{road network topology} that may be particularly pertinent to forecasting traffic flow. Similar motifs in the road network of the same region are likely to correspond to similar traffic dynamics---as such, automatically discovering and compressing them is likely to be useful. To enable such effects, we have directly adapted established unsupervised GNN methodologies such as graph auto-encoders \cite{kipf2016variational} and deep graph infomax \cite{velickovic2019deep}---for both of them, we demonstrate benefits to more accurate ETA predictions.

Additionally, different kinds of heuristical computations over road networks may require different ways of \emph{aggregating} information. For example, shortest path-finding would require \emph{optimising} the signals over a node's neighbourhood, whereas flow estimation may require \emph{summing} such signals. Inspired by the principal neighbourhood aggregation \cite{corso2020principal} architecture, we have investigated various combinations of GNN aggregation functions, showing them to be beneficial across many modelling scenarios.

\begin{figure*}[t]
\centering
    \includegraphics[width=0.3\linewidth]{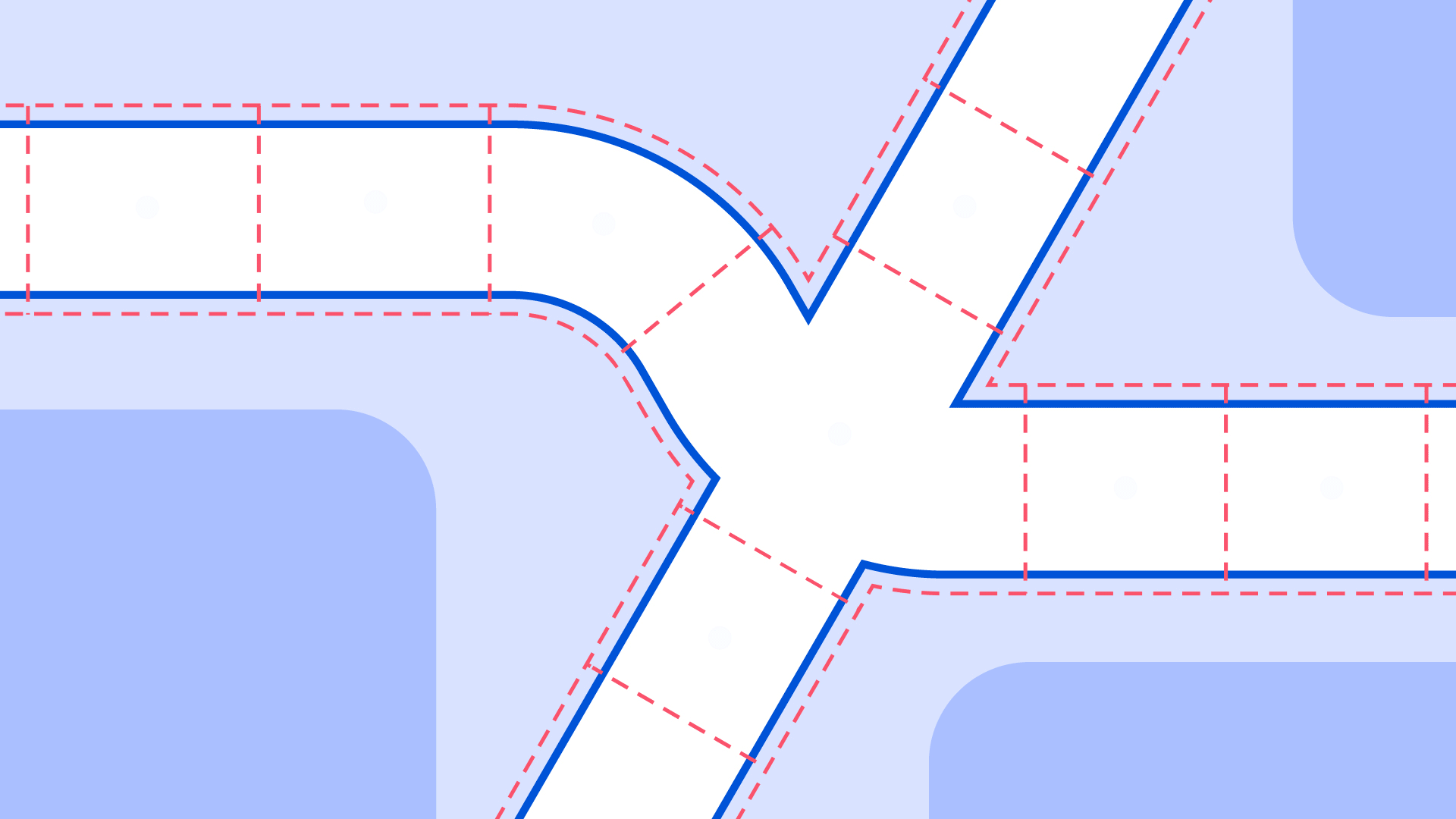} \hfill
    \includegraphics[width=0.3\linewidth]{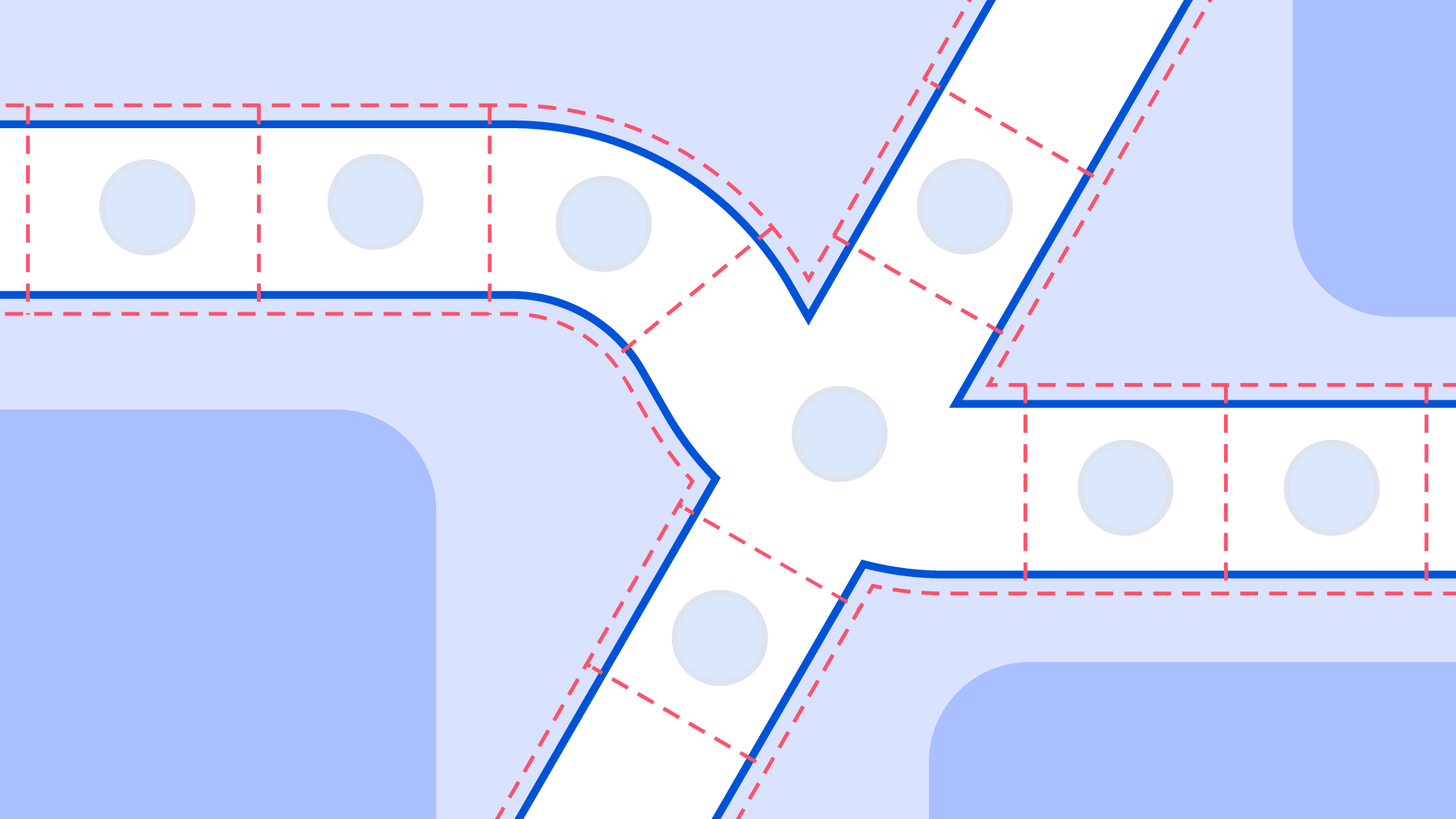} \hfill 
    \includegraphics[width=0.3\linewidth]{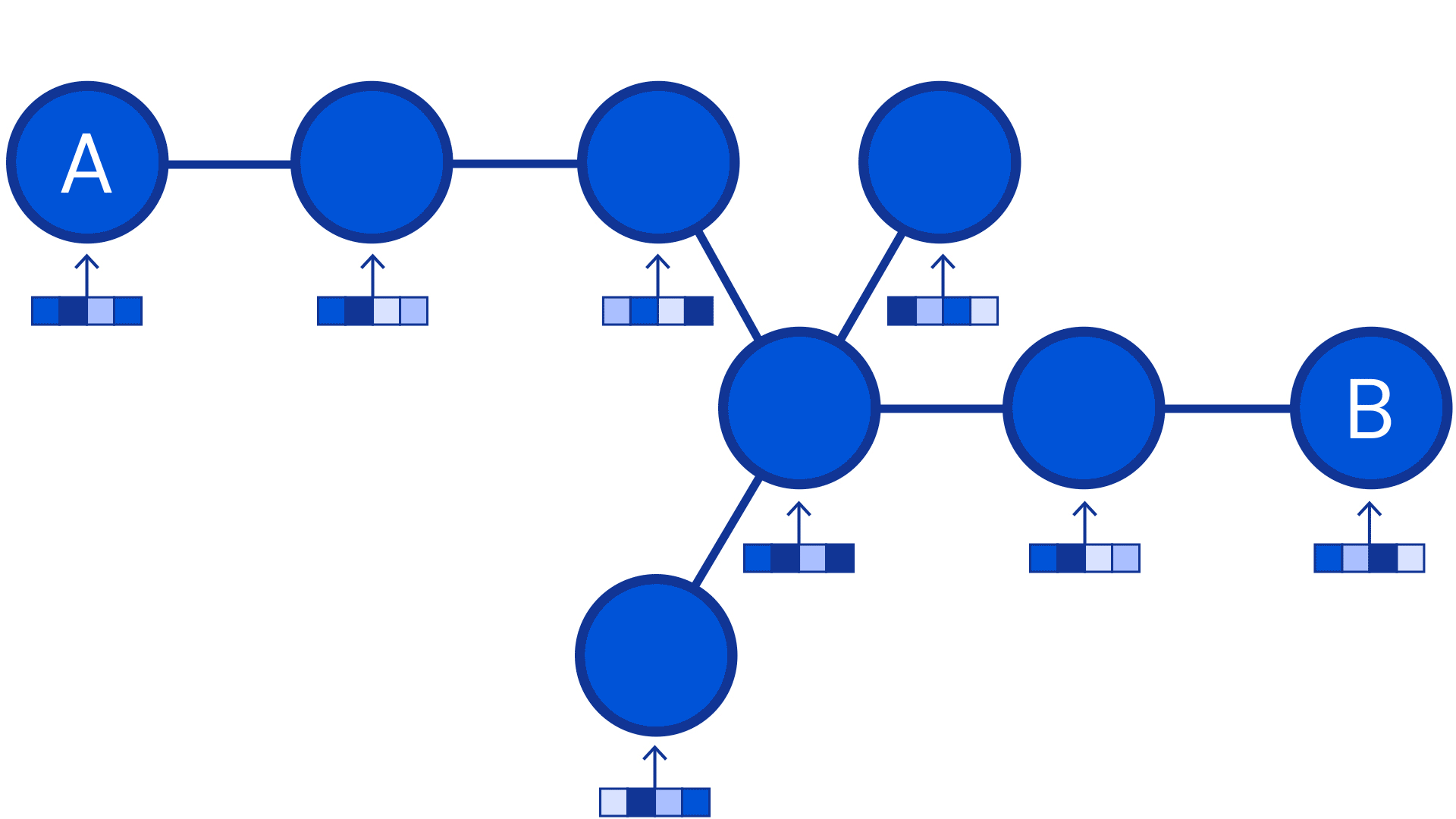}
\caption{An example road network with shared traffic volume, which is partitioned into segments of interest ({\bf left}). Each segment is treated as a node ({\bf middle}), with adjacent segments connected by edges, thus forming a \emph{supersegment} ({\bf right}). Note that for \emph{extended supersegments} (discussed in Section 4.1.4), extra off-route nodes may be connected to the graph.}
\label{fig:superseg}
\end{figure*}

\section{Method}
In this section, we describe our travel time prediction problem in the context of arrival time estimation. We also provide the technical details of the GNN models that we used for this problem, including their architecture and training details. 
The key components of our method are models that operate on networks of route segments in order to leverage their structure, and predict travel times for multiple time horizons into the future, in order to provide end users with more accurate ETAs. Key technical details of our GNN models are the various stabilizing methods that are used to reduce training variance, which have critical impact on user experience. Further, we will demonstrate that several changes to our launched model provide offline improvements under specific conditions.  

\subsection{Problem Setup} 
To achieve accurate travel time estimates, we model the road network using \textbf{supersegments}, which are sequences of connected road segments that follow typical traffic routes.
For a given starting time, we learn the travel time of each supersegment for different fixed time horizons into the future. At serving time, the sequence of supersegments that are in a proposed route are queried sequentially for increasing horizons into the future. This process specifically involves sequentially using the prediction time of an earlier supersegment to determine the relevant fixed horizons for the next supersegment, and interpolating between the prediction times of the fixed horizons to arrive at a travel time for the next supersegment. This process enables us to provide accurate travel time estimates in a scalable way, taking into account both spatial and temporal information.


A \emph{supersegment}, $\mathcal{S}$, is defined as a graph $\mathcal{S} = (S, E)$ where each node $s \in S$ is a road segment, and an edge $e_{ij} \in E$ exists if two segments $s_i$ and $s_j$ are connected (see Figure \ref{fig:superseg} for an overview). The supersegment travel time prediction task is then: given a supersegment $\mathcal{S}_t$, predict the travel time across the supersegment at different horizons into the future $y_t, y_{t+h_1}, y_{t+h_2}, ..., y_{t+h_k}$, where $h_1, h_2, ... h_k$ correspond to the different horizons.

\subsubsection{Data}
The world is divided into regions that confine similar driving behaviors and capture most trips without splitting. Data is constructed for each of these regions in order to build region-specific GNN models. For training and evaluation, each example was collected from a traversal event across a supersegment and its underlying segments. Specifically, a unique supersegment may appear in multiple examples corresponding to multiple traversals over that supersegment. The actual traversal times along segments and supersegments in seconds are used as node-level and graph-level labels for prediction, and each unique supersegment can appear in the training and evaluation data multiple times, corresponding to multiple traversals. The features, describing the traffic conditions of the supersegments prior to the traversal event, were collected backwards in time by the duration of each fixed horizon. The road segments are on average 50 - 100 meters long, and the supersegments contain on average about 20 road segments. Horizons we use are 0 seconds, 600 seconds, 1200 seconds, 1800 seconds, and 3600 seconds. In Section 4, we also investigate the trade-off of expanding supersegment graphs to include neighbouring nodes, or segments that extend beyond an immediate route.

\subsubsection{Features}
 For this task, we utilize both \emph{real-time} information that captures traffic patterns at the time of prediction and \emph{historical} data for traffic patterns specific to the time-of-day and day-of-week when the prediction is made. On a segment (node) level, we provide the average real-time and historical segment travel speeds and times, as well as segment length and segment priority (eg. road classifications such as highways) as features. On a supersegment (graph) level, we additionally provide real-time supersegment travel times as a feature. Real-time travel speeds and times are provided for 17 2-minute windows prior to horizon 0. Historical speeds and times are provided for five 8-minute windows prior to horizon 0, and seven 8-minute windows after horizon 0, with the values being an average across the past 17 weeks. We additionally provide learnable segment- and supersegment-level embedding vectors (of sizes 16 and 64, respectively). This enables sharing of information whenever the same segment appears in different supersegments, and whenever the same supersegment appears in different candidate routes.

\subsubsection{Baselines}
Possible baselines for this problem include \begin{inparaenum}
\item[(a)] \emph{nonparametric} methods, that compute travel times by averaging speeds across segments and supersegments.
\item[(b)] \emph{segment-level} models, that bypass the graph structure induced by supersegments. Such models predict travel times along every segment in isolation, adding them together to predict across supersegments. Prior production models used segment-level linear regression models.
\end{inparaenum}

\subsection{Model Architecture}

Our GNN model leverages full \emph{Graph Network} (GN) blocks exactly as described in \cite{battaglia2018relational}. Specifically, a GN block is defined with three ``update'' functions $\phi$ corresponding to edge, node, and global or supersegment-level updates, and three ``aggregation'' functions, $\rho$:
\begin{align}
  \begin{split}
    \mathbf{e}'_k &= \phi^e\left(\mathbf{e}_k, \mathbf{v}_{s_k}, \mathbf{v}_{t_k}, \mathbf{u} \right) \\
    \mathbf{v}'_i &= \phi^v\left(\mathbf{\bar{e}}'_i, \mathbf{v}_i, \mathbf{u}\right) \\
    \mathbf{u}' &= \phi^u\left(\mathbf{\bar{e}}', \mathbf{\bar{v}}', \mathbf{u}\right)
  \end{split}
  \begin{split}
    \mathbf{\bar{e}}'_i &= \rho^{e \rightarrow v}\left(E'_i\right) \\
    \mathbf{\bar{e}}' &= \rho^{e \rightarrow u}\left(E'\right) \\
    \mathbf{\bar{v}}' &= \rho^{v \rightarrow u}\left(V'\right)   
  \end{split}
  \label{eq:gn-functions}.
\end{align}
For each edge $k$, $\phi^e$ takes in edge features $e_k$, source and target node features $v_{s_k}$ and $v_{t_k}$, supersegment features $u$, and outputs an edge representation $e'_k$. For each node $i$, $\phi^v$ takes in node features $v_i$, aggregate edge representations of node $i$'s edges $\bar{e}'_i$, supersegment features $u$, and outputs a node representation $v'$. Finally, $\phi^u$ takes in aggregate node representations $\bar{v}'$, aggregate edge representations $\bar{e}'$, supersegment features $u$, and outputs a supersegment representation $u'$.


We compose 3 GN blocks into an \emph{encode-process-decode} architecture \cite{hamrick2018relational}, and learn a separate model for each horizon $h$. These blocks are defined as $\text{GN}_{enc}^{(h)}$, $\text{GN}_{proc} ^{(h)}$, $\text{GN}_{dec}^{(h)}$.

The encoder is first applied to the supersegment $\mathcal{S}$ with the previously described raw features. It produces \emph{latent} representations of nodes, edges and the supersegment itself. These latents are then passed to the processor, which is applied $2$ times (with shared parameters) to \emph{update} these representations. Finally, these representations are transformed into appropriate \emph{predictions} by applying the decoder.

The model predictions are $\hat{y}_{t+h}^{(i)}, \hat{y}_{j, t+h}^{(i)}, \hat{y}_{c, j, t+h}^{(i)}$, which are supersegment, segment, and cumulative segment predictions. On the supersegment level, the models predict the estimated travel time for the entire supersegment, $\hat{y}_{t+h}^{(i)}$. In addition, for every node, the models predict the estimated travel time across the segment $\hat{y}_{j, t+h}^{(i)}$ as well as the cumulative estimated travel time up to and including that segment $\hat{y}_{c, j, t+h}^{(i)}$, which we found are helpful for representation learning. In production, we use the supersegment-level output $\hat{y}_{t+h}^{(i)}$ instead of any of the node-level outputs. Intuitively, summing the individual segment predictions, for example, can lead to an accumulation of error.

The update functions, $\phi$, are all simple multilayer perceptrons (MLPs). Aggregation functions $\rho$ are \emph{summations}. While this is common practice, there exist benefits from using multiple aggregator functions like \emph{summing} and \emph{averaging} and concatenating their results together \cite{corso2020principal}. For certain regions and horizons in offline testing, this was true in our case as well. We will discuss the comparative benefits of various aggregators in Section 4.

\subsection{Model Training}
\begin{table*}[h!]
\centering
\begin{tabular}{c c c c c c}
    \toprule
    Dataset & \shortstack{\# (Unique) Training \\ Supsersegments} & \shortstack{\# (Unique) Testing \\ Supersegments} & \shortstack{Average \# Segments \\ per Supersegment} & \shortstack{Average \# Segments \\ per Extended Supersegment} & \shortstack{Average Segment \\ Length (m)}\\
    \midrule
    LAX & 661M (16k) & 126M (16k) & 21.79 & 204.60 & 81.02 \\
    NYC & 502M (12k) & 91M (12k) & 20.75 & 147.67 & 109.03 \\
    SGP & 257M (5k) & 54M (5k) & 18.80 & 96.31 & 106.29 \\
    TYO & 115M (4k) & 22M (4k) & 24.52 & 236.54 & 57.94 \\
    \bottomrule
\end{tabular}
\caption{Information on datasets, including the number of total supersegments, and unique supersegments in each training and evaluation dataset, the average number of segments in a supersegment computed across a training dataset, and the average length of segments computed across a training dataset. }
\label{table:datasets}
\end{table*}
While training our GNN models, we found it very useful to use a \emph{combination} of various loss functions. This resulted in strong serving performance when deployed in production. The primary bottleneck to deploying was the \emph{high variability} of models during training. Variance is an issue in application because models saved at different points of training will result in different end user experiences. To address this, we use a MetaOptimizer that makes use of \textbf{MetaGradients} \cite{xu2018meta} for adapting the learning rate during training, as well as applying an exponential moving average over model parameters during evaluation and serving.

\subsubsection{Losses}
Although our main objective is to predict supersegment-level travel time, we found that using auxiliary losses for good segment-level travel time prediction helps the final performance. Our final loss for a specific horizon $h$ is
\begin{equation}\label{eqn:losses}
 L = \ell_{\text{ss}} + \lambda_{\text{s}}\ell_{\text{s}} + \lambda_{\text{sc}}\ell_{\text{sc}},
\end{equation}
with additional weight decay, where $ \lambda_{\text{s}} = 1$, $ \lambda_{\text{sc}} = 0.15$ and
\begin{align}
\ell_\text{ss} &= \sum_{i = 0}^{N} \Bigg(\frac{1}{\max\{f^{(i)}, 1\}^{0.75}} \Bigg) \cdot \mathcal{L}_{400}\big(y_{t+h}^{(i)}, \hat{y}_{t+h}^{(i)}\big)\\
\ell_\text{s} &= \sum_{i = 0}^{N} \sum_{j = 0}^{m^{(i)}} \Bigg(\frac{1}{\max\{f_j^{(i)}, 1\}^{0.75}}\Bigg) \cdot \mathcal{L}_{400}\big(y_{j, t+h}^{(i)}, \hat{y}_{j, t+h}^{(i)}\big)\\
\ell_\text{sc} &= \sum_{i = 0}^{N} \sum_{j = 0}^{m^{(i)}} \Bigg(\frac{1}{\max\{\sum_{k = 0}^{j} f_k^{(i)}, 1\}^{0.75}}\Bigg) \cdot \mathcal{L}_{400}\big(\sum_{k = 0}^{j} y_{k, t+h}^{(i)}, \hat{y}_{c, j, t+h}^{(i)}\big).
\end{align}
The provided labels and weights are $y_{t+h}^{(i)}$ and $y_{j, t+h}^{(i)}$, which are supersegment and segment traversal times, and $f^{(i)}$ and $f_{j}^{(i)}$, which are supersegment and segment free flow times, an estimate of traversal times when little to no traffic is present. We predict $\hat{y}_{t+h}^{(i)}$ in addition to segment-level travel times in order to avoid an accumulation of errors over segments.
$\mathcal{L}_{\delta}$ is a Huber loss function with delta value $\delta$, which is important in being less sensitive to outliers. The losses are summed up across the segments and supersegments, so losses such as the supersegment-level loss is more influenced by supersegment length whereas losses like the segment-level loss is more independent from length. We exponentially scale the Huber loss down for examples with greater free flow times to better control the loss's growth.

We also experimented with purely \emph{unsupervised} auxiliary losses, and will discuss their trade-offs in Section 4.

\subsubsection{MetaGradients}

In order to reduce the variance of the recovered GNN models across epochs, we resort to a careful treatment of the \emph{learning rate} for each step of training. Our method of choice, MetaGradients \cite{xu2018meta}, follows the online cross-validation paradigm and jointly optimizes the hyper-parameter $\eta$ with the model parameters $\theta$ at each training iteration. Although the original work targets at reinforcement learning, the approach can be generalized well to supervised learning applications. MetaGradients frames the optimization problem as $L(\tau, \theta, \eta)$, where $\tau$ is a training example, $\theta$ are the model parameters, $\eta$ is the hyper-parameter (e.g., the learning rate) and $L$ is the original loss function additionally parameterized by $\eta$. For a new training example, $\tau'$, an underlying update to the model parameters $\theta' = \theta + f(\tau, \theta, \eta)$ with some update function $f$ (eg. a gradient step) additionally parameterized by $\eta$, and a reference meta-parameter value $\eta'$, MetaGradients computes the update for $\eta$ using the following derivative:
\begin{equation}
\frac{\partial L(\tau', \theta', \eta')}{\partial\eta
} = \frac{\partial L(\tau', \theta', \eta')}{\partial\theta'}\frac{\partial\theta'}{\partial\eta}~,
\end{equation}
where the gradient w.r.t. the model parameter is multiplied by a factor $\partial\theta'/\partial\eta$. Xu et al. \cite{xu2018meta} has shown that, in practice, the second term can be approximated using an accumulative trace, i.e.,
\begin{equation}
    \frac{\partial\theta'}{\partial\eta}=\frac{\partial\theta}{\partial\eta}+\frac{\partial L(\tau, \theta, \eta)}{\partial\eta
}~.
\end{equation}
In our application, we have successfully combined MetaGradients with the Adam optimizer \cite{kingma2014adam}.

\subsubsection{Exponential Moving Average (EMA) of Parameters} During evaluation and serving, we use an EMA of model parameters 
\begin{equation}
\theta_{\text{EMA}} = \alpha \cdot \theta_{\text{EMA}} + (1 - \alpha) \cdot \theta,
\end{equation}
where $\alpha = 0.99$. EMAs were another key component in reducing variance in saved models within training runs.

\section{Experiments}

\subsection{Experimental Setup}
\begin{table*}[h!]
\centering
\begin{tabular}{c c c c c c c c c}
\toprule
\multirow{2}{*} & \multicolumn{2}{c}{\bf Real-Time} & \multicolumn{2}{c}{\bf Historical} & \multicolumn{2}{c}{\bf DeepSets} & \multicolumn{2}{c}{\bf GN} \\
 & Horizon 0 & Horizon 3600 & Horizon 0 & Horizon 3600 & Horizon 0 & Horizon 3600 & Horizon 0 & Horizon 3600 \\
\toprule
NYC & 42.31 & 69.60 & 48.03 & 51.80 & 37.33 & 46.68 & \textbf{36.84} & \textbf{46.55} \\
LAX & 47.58 & 79.31 & 53.40 & 60.22 & 40.71 & 50.06 & \textbf{40.32} & \textbf{49.94} \\
TYO & 63.56 & 81.01 & 63.76 & 67.63 & 52.15 & 60.07 & \textbf{51.64} & \textbf{59.86} \\
SGP & 45.09 & 75.83 & 58.25 & 61.39 & 37.56 & 50.81 & \textbf{36.79} & \textbf{50.67} \\
\bottomrule  
\end{tabular}
\caption{Offline model performance against baseline methods reported in RMSE, averaged across five random seeds. The graph network (GN) significantly outperforms all reported baselines for each region and horizon ($p < 0.01$ using a $t$-test).}
\label{table:baselines}
\end{table*}
\subsubsection{Datasets}
Although our models were launched across regions world-wide, we evaluate our model specifically over: \emph{New York} ({\bf NYC}), \emph{Los Angeles} ({\bf LAX}), \emph{Tokyo} ({\bf TYO}) and \emph{Singapore} ({\bf SGP}), finding that the results generalize to other regions (see Figure \ref{fig:maps_impro}). For offline evaluation, we used training data collected during January 2020. For online evaluation, our data was collected during November 2020. We note that November 2020 data may be susceptible to COVID-19 traffic patterns, and hence we attempted to select regions that minimize this deviation for evaluation in order to capture performance under regular traffic conditions. The distributions of the datasets and their extended versions are described in Table \ref{table:datasets}.

\subsubsection{Baselines}
We evaluate our GNN models against nonparametric travel times computed using real-time speeds and historical speeds, as well as models that do not leverage graph structure:
\begin{itemize}
    \item \textbf{Real-time travel times}: Summation of segment-level travel times computed using segment speeds averaged over a two minute window prior to the time of prediction. These are numbers computed from data.
    \item \textbf{Historical travel times}: Summation of segment-level travel times computed using segment speeds specific to hour of day and day of week at the time of prediction, averaged across last 17 weeks. These are numbers computed from data.
    \item \textbf{DeepSets}: Transform node representations using a segment-level feed-forward network followed by a sum aggregation over the segment-level representations before using a final feed-forward network to make graph-level predictions \cite{zaheer2017deepsets}. This effectively treats supersegments as a ``bag-of-segments'', \emph{ignoring} their graph structure.
\end{itemize}
\subsubsection{Ablations of Launched Models}
We ablate over the specific elements of our launched GNNs that contribute to performance improvements and variance reduction. This includes learnable segment and supersegment embeddings.

\paragraph{Embeddings}
We ablated over the embeddings defined for unique segment and supersegment IDs. We use a dimension of $16$ for segment-level embeddings and a dimension of $64$ for supsersegment-level embeddings across all models that use embeddings. The embedding vocabularies are region-specific, covering $99.5$\% of segment IDs and supersegment IDs in the corresponding datasets, and assigning the rest to be out-of-vocabulary (OOV), spread out across $200$ and $20$ OOV buckets for segments and supersegments respectively.

\paragraph{MetaGradients \& EMA}
We directly compare our models trained with and without MetaGradients. When trained with MetaGradients, we set the meta-learning rate to $0.01$ and the update frequency to every $100$ iterations.

We compare our GNN models that use EMA (with a decay factor of $\alpha = 0.99$) to models that do not.

\subsubsection{Notable Extensions}
We describe and compare extensions to the launched GNN models that show offline improvements, and can be useful in specific settings. This includes extended supersegment graph data, combinations of aggregators, and unsupervised auxiliary losses.

\paragraph{Extended Supersegments}
We investigate performance improvements for GNN models that operate on \textbf{extended} supersegments. Extended supersegments are larger graphs that include nodes from neighbouring segments, in addition to the original segments. These include segments from nearby, possibly connected traffic. The prediction task and labels remain the same. Specifically, the goal is still to predict travel times across the original segments and supersegments within the extended supersegments.

This enables our models to make use of additional spatial context---such as traffic congestion on a connecting route---at the expense of data storage, slower training, and inference. We note that the same GNN architecture is used for both supersegment and extended supersegment experiments, making the \emph{model} storage cost remain constant. Further, we experiment with binary features that indicate whether segments are on the main route in an extended graph. 

\paragraph{Combinations of Aggregators}
We further investigated the benefits of swapping out the default \emph{sum} aggregation function with \emph{combinations} of different aggregators. We combine by \emph{concatenating}; for example, the combination $[\text{Min}, \text{Max}]$ has an aggregation function for nodes of:
\begin{equation} \rho^{s\rightarrow u}_i(S) =  \left[\min_{j\in N(i)} s_j || \max_{j\in N(i)} s_j \right] \end{equation}
where $N(i)$ are the road segments that neighbour $s_i$. We ablate over all combinations of Max, Min, SqrtN, and Sum. In the interests of brevity, only the best-performing ones are included in Table ~\ref{table:aggregators-updated}.

\paragraph{Unsupervised Auxiliary Losses}
Lastly, we experimented with incorporating \emph{unsupervised} losses, which are designed to produce representations that capture the \emph{structural} information of input graphs. This, in turn, may simplify discovery of interesting motifs within the road network topology, which may implicitly drive traffic dynamics. In all experiments, these losses are combined with the supervised objective (Equation \ref{eqn:losses}) to assist learning. 

We combine two main kinds of unsupervised loss functions. Firstly, \emph{Deep graph infomax} (DGI) \cite{velickovic2019deep} encourages the node embeddings to encode \emph{global} structural properties, by carefully contrasting the road network against an artificially \emph{corrupted} version. Conversely, the \emph{graph auto-encoder} (GAE) \cite{kipf2016variational} leverages a \emph{link prediction}-style loss and is hence mindful of \emph{local} topology.

\subsection{Offline Evaluation}
\begin{table*}[h!]
\centering
\begin{tabular}{c c c c c c c c c}
\toprule
\multirow{2}{*} & \multicolumn{2}{c}{\bf GN} & \multicolumn{2}{c}{\bf \shortstack{GN without\\ Segment Embeddings}} & \multicolumn{2}{c}{\bf \shortstack{GN without\\ Supersegment Embeddings}} & \multicolumn{2}{c}{\bf \shortstack{GN without Segment \underline{and}\\ Supersegment Embeddings}} \\
 & Horizon 0 & Horizon 3600 & Horizon 0 & Horizon 3600 & Horizon 0 & Horizon 3600 & Horizon 0 & Horizon 3600 \\
\midrule
NYC & \textbf{36.87} & \textbf{46.59} & 37.03 & 46.59 & 37.04 & 46.61 & 37.55 & 46.86 \\
LAX & 40.39 & \textbf{49.96} & 40.48 & 50.03 & \textbf{40.38} & 49.96 & 41.11 & 50.33 \\
TYO & 51.76 & \textbf{59.90} & 52.02 & 59.98 & \textbf{51.72} & 60.00 & 52.63 & 60.60 \\
SGP & \textbf{36.84} & 50.60 & 37.29 & 50.83 & 36.98 & \textbf{50.52} & 37.41 & 51.25 \\
\bottomrule  
\end{tabular}
\caption{An ablation on the segment and supersegment learnable embedding representations used as input to the Graph Net model. Performance for each reported in RMSE.}
\label{table:embeddings}
\end{table*}
For offline evaluation, we use the root mean squared error (RMSE) between the predicted supersegment travel times and the actual traversal times across supersegments, in seconds:
\begin{equation}
\text{RMSE}_\text{ss} = \sqrt{\frac{1}{N}\sum_{i = 0}^{N} \big(y_{t+h}^{(i)} - \hat{y}_{t+h}^{(i)}\big)^2}.
\end{equation}
Reported values are computed using the temporally held-out test datasets of the same regions. Other metrics such as mean absolute error (MAE) show similar results and are thus omitted. RMSE is used to make decisions for which architectures are used in online evaluations and launches, so we focus on that metric here.

\paragraph{Baseline Comparison}
Table \ref{table:baselines} shows the offline evaluation results of our GNN models compared to the baselines, averaged over five random seeds. Across all studied regions and horizons, our GNN models outperform the baselines. Compared to the closest baseline, DeepSets, our GNN models show an improvement of 0.12 to 0.77 RMSE. One note is that these improvements are computed over individual supersegments, and they will accumulate over entire routes---hence our improvements may quickly become more important on a whole-route level. Across all of the considered regions and horizons, the improvements of the GNN are \emph{\textbf{statistically significant}}: $p < 0.01$ using a $t$-test on the individual run results, across five runs.

\paragraph{Embeddings}
Table \ref{table:embeddings} studies the effect of the various learnable input embeddings, by attempting to ablate them out. Across most regions and horizons, such embeddings are useful for improving RMSE.

\begin{figure*}[t]
\centering
    \includegraphics[width=0.45\linewidth]{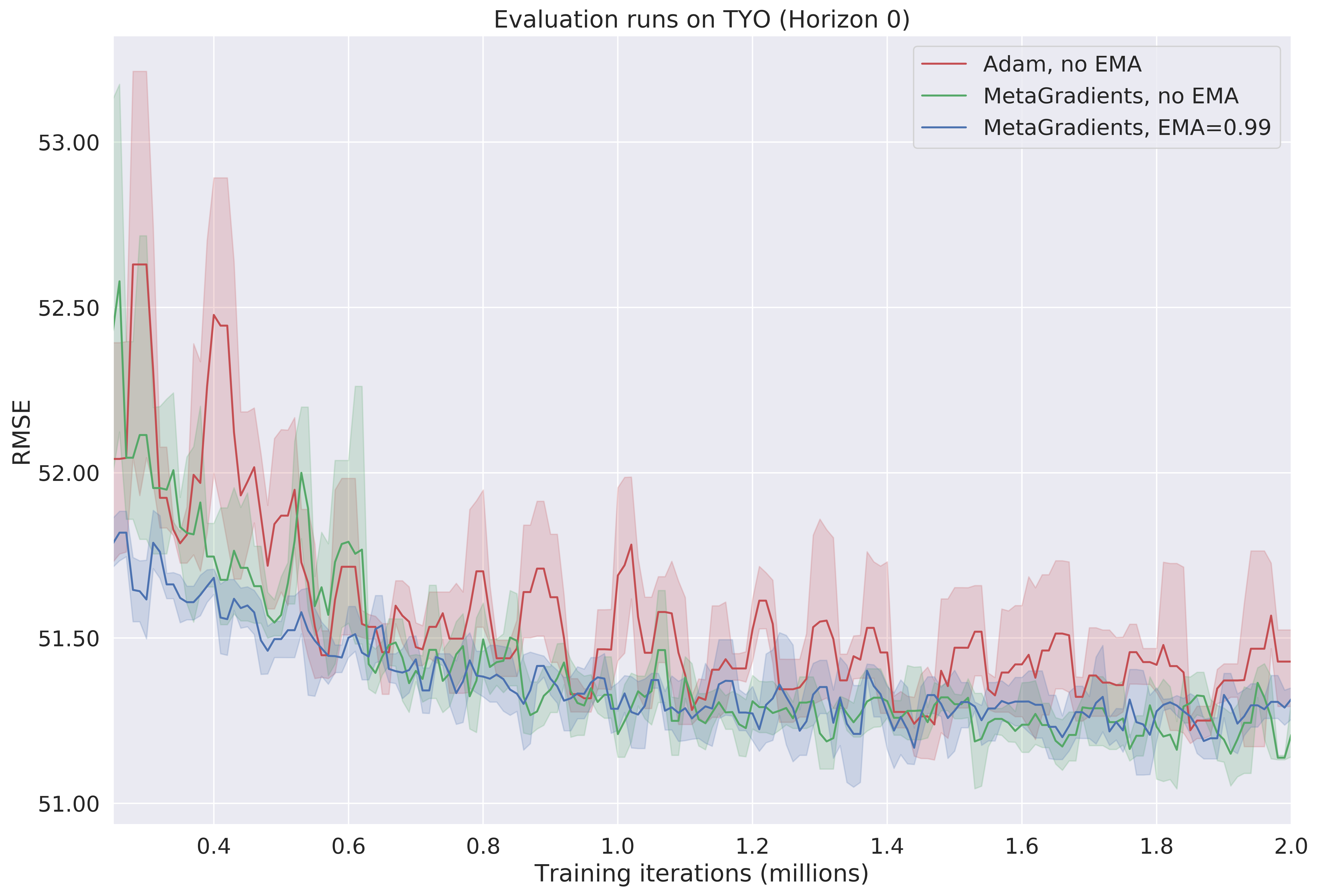} \hfill
     \includegraphics[width=0.45\linewidth]{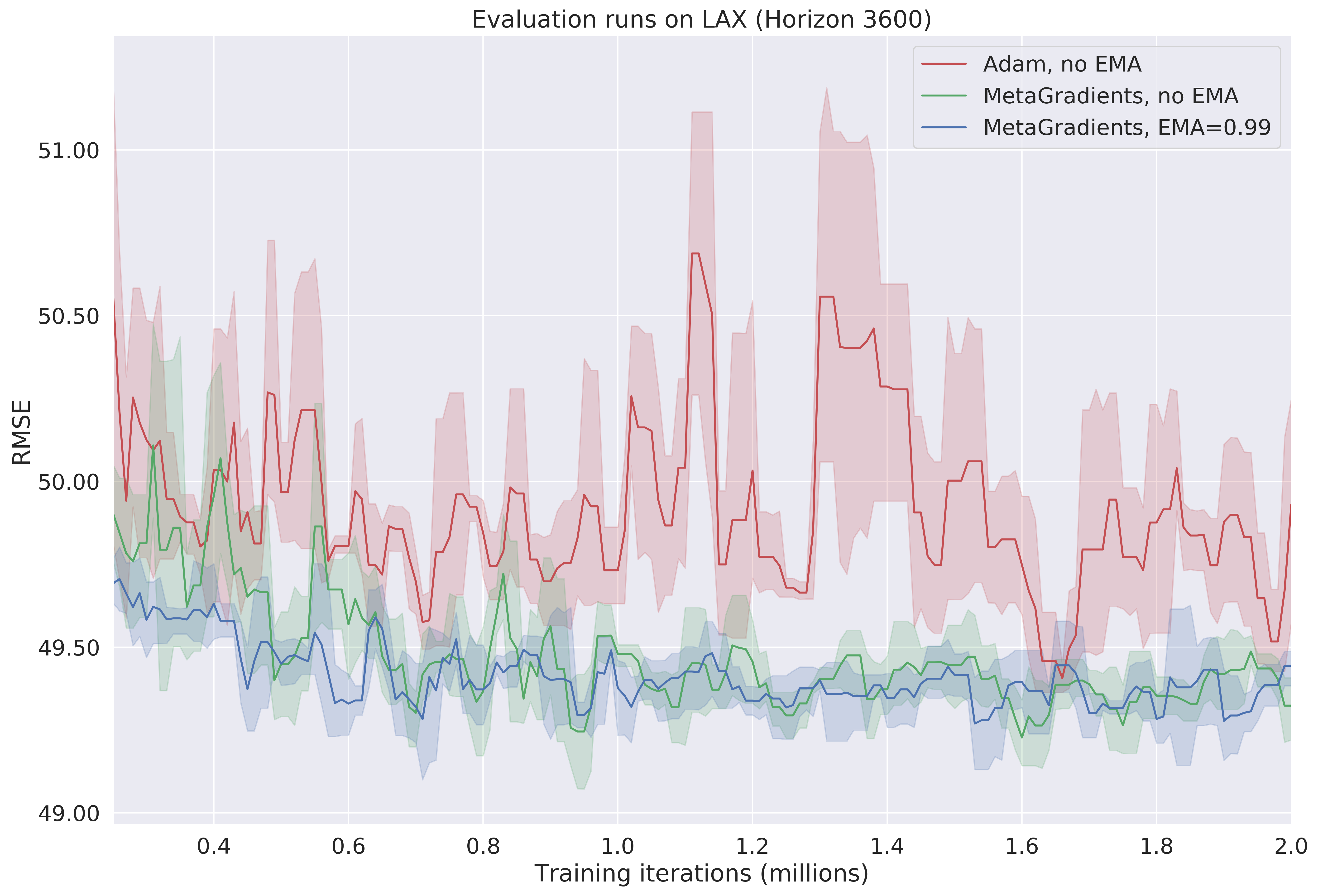}
\caption{Validation RMSE during training, with and without MetaGradients and EMA decay, aggregated across five seeds. Both methods contribute to variance reduction. Results shown for two metro/horizon setups -- same trends hold elsewhere.}
\label{fig:metaopt-ema}
\end{figure*}


\paragraph{MetaGradients \& EMA}
Figure \ref{fig:metaopt-ema} compares training with and without applying MetaGradients and EMA decay over several seeds. For brevity, we only show results for two settings (TYO Horizon 0, LAX Horizon 3600); the trends generalize across different regions and horizons. Both MetaGradients and EMA contribute to lowering within-run and across-run variance, which are critical for inference in production. Further, it was shown that MetaGradients consistently decays the learning rate over time, which ensures stable training.

\begin{table*}[h!]
\centering
\begin{tabular}{c c c c c c c}
\toprule
\multirow{2}{*} & \multicolumn{2}{c}{\bf Supersegments} & \multicolumn{2}{c}{\bf Extended Supersegments} & \multicolumn{2}{c}{\bf \shortstack{Extended Supersegments\\ + Extra Features}} \\
 & Horizon 0 & Horizon 3600 & Horizon 0 & Horizon 3600 & Horizon 0 & Horizon 3600 \\
\midrule
NYC & 35.85 & 45.91 & 36.11 & 45.71 & \textbf{35.64} & \textbf{45.65} \\
LAX & 39.51 & 49.01 & 39.67 & 48.99 & \textbf{39.22} & \textbf{48.79} \\
TYO & 52.59 & 60.72 & 53.24 & 60.85 & \textbf{52.57} & \textbf{60.64} \\
SGP & 34.89 & 49.51 & 35.29 & \textbf{49.48} & \textbf{34.73} & 49.50 \\
\bottomrule  
\end{tabular}
\caption{A comparison between different input graph formats. Travel time predictions only include road segments in the traversed track. Same embedding vocabularies are used across variants. Performance for each reported in RMSE.}
\label{table:extended}
\end{table*}

\paragraph{Extended Supersegments}
Table \ref{table:extended} compares GNNs that operate on supersegments to variants that operate on extended supersegments. Combining extended supersegments with additional binary features that indicate whether a segment or an edge is part of the original supersegment is capable of improving RMSE (by 0.01 to 0.29). In a scenario where data storage and inference cost increases are not core issues, the extended supersegments may hence be a useful data representation.


\begin{table*}[h!]
\centering
\begin{tabular}{c c c c c c c c c}
\toprule
\multirow{2}{*} & \multicolumn{2}{c}{\bf NYC} & \multicolumn{2}{c}{\bf LAX} & \multicolumn{2}{c}{\bf TYO} & \multicolumn{2}{c}{\bf SGP} \\
 & Horizon 0 & Horizon 3600 & Horizon 0 & Horizon 3600 & Horizon 0 & Horizon 3600 & Horizon 0 & Horizon 3600 \\
\midrule
GN & 36.88 & 46.56 & 40.30 & 49.95 & 51.59 & 59.82 & 36.91 & 50.77 \\
GN+DGI & 36.79 & 46.53 & \textbf{40.18} & \textbf{49.90} & 51.48 & \textbf{59.76} & \textbf{36.76} & \textbf{50.54} \\
GN+GAE & \textbf{36.76} & \textbf{46.38} & 40.23 & 49.91 & \textbf{51.43} & 59.82 & 36.85 & 50.57 \\
\bottomrule
\end{tabular}
\caption{Compares using different unsupervised algorithms for augmenting the Graph Net as an auxiliary loss function. Performance for each reported in RMSE.}
\label{table:unsupervised-updated}
\end{table*}


\paragraph{Unsupervised Auxiliary Losses}
Table ~\ref{table:unsupervised-updated} compares the performance between different unsupervised losses in terms of RMSE. Augmenting the loss function with deep graph infomax (DGI) or graph auto-encoders (GAE) by a tuned weight parameter achieves an improvement of 0.05 to 0.23 RMSE. The optimal variant and hyperparameters were different for each region and prediction horizon. Thus, it is likely expensive to tune such losses for all existing regions and each horizon. However, GN+DGI would be a reasonable option as it showed some degree of improvement in all cases. Note that there are several important design decisions involved with deploying DGI: e.g. deciding the readout, corruption, and discriminator functions. We, however, did not explore many options in this space, as the potential performance improvements were not deemed significant enough for the additional costs incurred in a production environment.


\begin{table*}[h!]
\centering
\begin{tabular}{c c c c c c c c c}
\toprule
\multirow{2}{*} & \multicolumn{2}{c}{\bf NYC} & \multicolumn{2}{c}{\bf LAX} & \multicolumn{2}{c}{\bf TYO} & \multicolumn{2}{c}{\bf SGP} \\
 & Horizon 0 & Horizon 3600 & Horizon 0 & Horizon 3600 & Horizon 0 & Horizon 3600 & Horizon 0 & Horizon 3600 \\
\midrule
Sum & 36.88 & 46.56 & \textbf{40.30} & 49.95 & \textbf{51.59} & \textbf{59.82} & 36.91 & 50.77 \\
SqrtN & 37.06 & 46.60 & 40.39 & 49.95 & 51.75 & 59.96 & 36.93 & 50.66 \\
Mean & 36.92 & 46.60 & 40.45 & 49.98 & 51.92 & 60.01 & \textbf{36.78} & 50.66 \\
Min & 36.98 & 46.57 & 40.49 & 49.91 & 51.83 & 59.95 & 36.98 & 50.81 \\
Max & 37.01 & 46.57 & 40.48 & 49.95 & 51.84 & 59.96 & 36.97 & 50.94 \\
All & \textbf{36.83} & \textbf{46.53} & 40.33 & \textbf{49.88} & 51.60 & 59.90 & 36.95 & \textbf{50.43} \\
\bottomrule
\end{tabular}
\caption{Performance in RMSE using various aggregation functions. Only the individual aggregation operations and the full combination are included for brevity due to similar or worse results. Note that the optimal aggregation combination differs between regions as well as horizons.}
\label{table:aggregators-updated}
\end{table*}


\paragraph{Combinations of Aggregators}
Table \ref{table:aggregators-updated} compares model performance when using different aggregation functions. When compared to our production default of sum, other variants may achieve an improvement of up to 0.34 RMSE depending on the region and horizon. Since the optimal aggregation function differs between regions and horizons, it is likely difficult and expensive to tune for all regions and horizons of the world. Using all five aggregations at once is a reasonable preference, and it may be beneficial to investigate dynamic aggregators as a future line of work. 

\subsection{Online Evaluation}
We compare our GNNs against baselines in an online setting in order to show the effectiveness of the models during serving. Both real-time travel times and historical travel times are used in precursor systems and fall-back systems. For online evaluation, we use:
\begin{equation}
\text{RMSE}_\text{track} = \sqrt{\frac{1}{N}\sum_{i = 0}^{N} \big(u^{(i)} - \hat{u}^{(i)}\big)^2}
\end{equation}
which is the root mean squared error between the predicted track travel times and actual travel times, computed over all tracks within a week, specifically January 15th 6:00 PM - January 22nd 2021 6:00 PM. We note that the GN and DeepSets models additionally use a series of fallback models for segments that do not appear in supersegments, and segments that do not have real-time or historical travel time input features, whereas the same is not done for real-time and historical travel times. 


\begin{table}[t]
\centering
\begin{tabular}{ c  c c c c }
    \toprule
    Region & Real-Time & Historical & DeepSets & GN \\
    \midrule
    NYC & 130.05 & 154.83 & 119.76 & \textbf{118.63} \\
    LAX & 154.19 & 171.25 & 139.02 & \textbf{138.10} \\
    TYO & 211.85 & 250.30 & 182.73 & \textbf{181.98} \\
    SGP & 127.54 & 138.59 & 109.39 & \textbf{108.78} \\
    \bottomrule
\end{tabular}
\caption{Online model performance against baselines reported in RMSE of track travel times.}
\label{table:online}
\end{table}

Table \ref{table:online} shows the results of our GNN models compared to the baselines in the online setting. It shows a similar pattern compared to Table \ref{table:baselines} with GNNs outperforming the baseline methods. Compared to DeepSets, the strongest baseline approach, GNNs outperform by 0.05 to 1.35 RMSE in this online setting.

\subsection{Analysis}
\begin{figure*}[t]
    \centering
    \includegraphics[width=\linewidth]{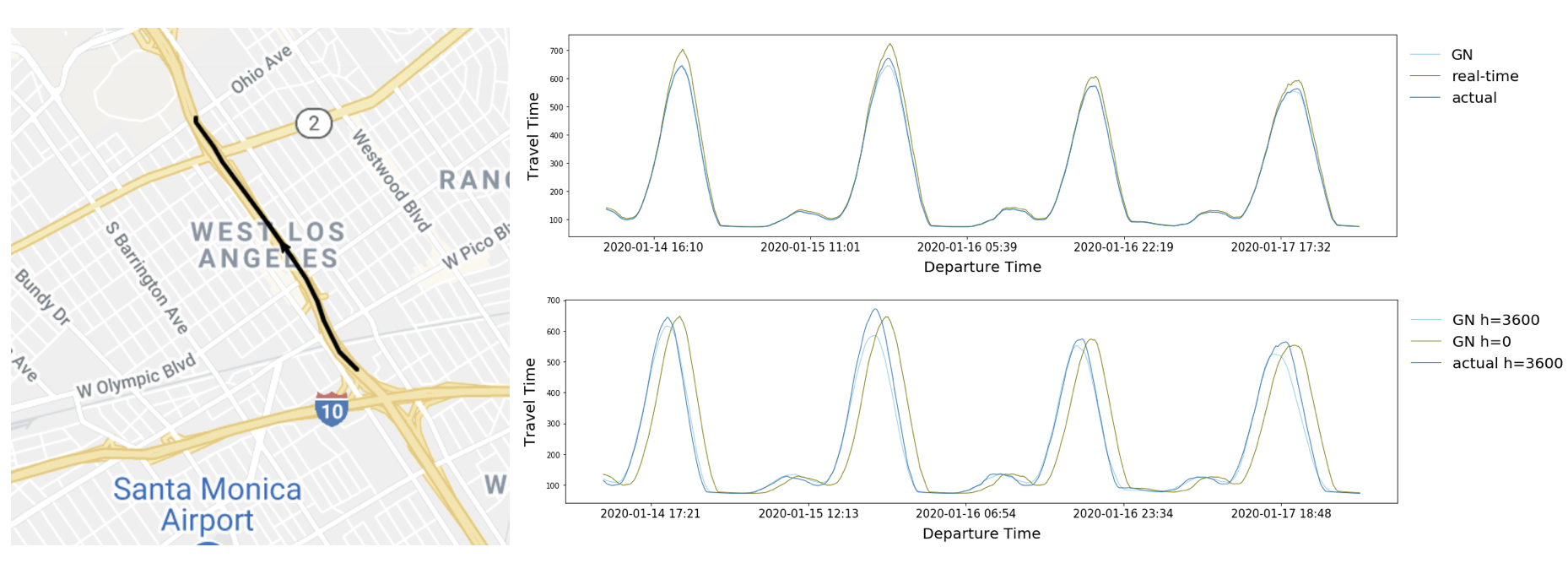}
    \caption{Qualitative analysis over a supersegment in LAX (shown on the left). Above: Travel times and predictions over different departure times. Our GNN is able to detect traffic congestions more accurately than using real-time travel times. Below: Travel times and predictions for different horizons into the future, over different route departure times. Having multiple prediction horizons is important to providing accurate ETA.}
    \label{fig:ss_map}
\end{figure*}

In this section, we provide specific examples that provide insight into scenarios where our GNN models are able to provide more accurate estimated time of arrivals compared to baselines, by better making use of spatial and temporal information. Figure \ref{fig:ss_map} (left) shows the supersegment (which is part of LAX) which we will use to illustrate these qualitative findings.



Figure \ref{fig:ss_map} (above) compares the predicted travel times over the supersegment to the actual travel times, over a range of departure times. The upward slopes of the GNN predicted travel times match the upward slopes of the actual travel times better than the real-time travel times baseline. In this setting, our GNN model is able to detect traffic congestions more accurately than the baseline by making use of traffic information from a larger neighborhood.

Figure \ref{fig:ss_map} (below) compares predicted travel times over the supersegment for various horizons into the future. The predictions of our GNN for $h=3600$ better match the actual travel times 60 minutes into the future, compared to the predictions of our GNN for $h=0$. This example illustrates the value of having multiple prediction horizons in providing accurate ETA.

\section{Engineering Challenges}
\textbf{Caching} It is challenging to meet the latency requirements of Google Maps while keeping the cost of evaluating Graph Net models low. The path in an ETA request usually includes multiple supersegments and making predictions for these supersegments on the fly is not practical nor scalable. Approaches like CompactETA \cite{fu2020compact} learn high level representations that can be queried during inference time for use in a very simple MLP. To resolve this issue, we instead created a shared lookup table caching fresh predictions for predefined supersegments for a fixed set of horizons, which is periodically updated. The server fetches necessary predictions upon a request and interpolates the predictions of adjacent horizons to compute the estimated arrival time for each supersegment. We have tested an update frequency of 15 seconds and found no measurable regression up to a frequency of 2 minutes, indicating staleness is not an issue.
\newline
\newline
\textbf{Turn speeds} It is common to have a multimodal speed distribution for a busy road segment heading to a split of traffic, such as an intersection or a highway ramp. For such cases, we used the real time and historical speed for the specific turn which the supersegment follows.
\newline
\newline
\textbf{Coverages} Currently we have about 1 million predefined supersegments that cover the most common routes taken by our users. The effect is that most freeways, major arterials and some popular short-cuts in metropolitan areas are selected, whereas less busy surface streets will not be covered. Multiple supersegments may cover any given road segments to account for multiple common routes that traverse through that road segment. For less frequently visited road segments, we use simpler per-segment models.

\section{Conclusion}
We presented how a graph neural network can be engineered to accurately predict travel time estimates along candidate routes. Applying stabilising techniques such as MetaGradients and EMA was a necessary addition to make the GNNs production-ready. We deployed our GNN model for ETA prediction in Google Maps, where it now serves user queries worldwide. We presented offline metrics, online evaluations, and user studies: all showing significant quantitative improvements of using GNNs for ETA predictions. Further, through extensive ablations on various design and featurization choices, we provide prescriptive advice on deploying these kinds of models in practice. We believe our findings will be beneficial to researchers applying GNNs to any kind of transportation network analysis---which is certain to remain a very important application area for graph representation learning as a whole.

\begin{acks}
We would like to thank Paulo Estriga and Adam Cain for designing several of the figures featured in this paper. 
\end{acks}

\bibliographystyle{ACM-Reference-Format}
\bibliography{sample-sigconf}



\end{document}